\documentclass[sigconf]{acmart}
\usepackage{amsmath}
\usepackage{multirow}
\usepackage{url}
\usepackage{enumitem}
\usepackage{dsfont}
\AtBeginDocument{%
  \providecommand\BibTeX{{%
    \normalfont B\kern-0.5em{\scshape i\kern-0.25em b}\kern-0.8em\TeX}}}

\begin{document}

\title{A Deep Graph Reinforcement Learning Model for Improving User Experience in Live Video Streaming}

\author{Stefanos Antaris}
\affiliation{
	\institution{KTH Royal Institute of Technology \\ Hive Streaming AB}
  	\country{Sweden}
}
\email{antaris@kth.se}

\author{Dimitrios Rafailidis}
\affiliation{%
  \institution{University of Thessaly}
  \country{Greece}}
\email{draf@uth.gr}

\author{Sarunas Girdzijauskas}
\affiliation{%
  \institution{KTH Royal Institute of Technology}
  \country{Sweden}
}
\email{sarunasg@kth.se}



\begin{abstract}
In this paper we present a deep graph reinforcement learning model to predict and improve the user experience during a live video streaming event, orchestrated by an agent/tracker. We first formulate the user experience prediction problem as a classification task, accounting for the fact that most of the viewers at the beginning of an event have poor quality of experience due to low-bandwidth connections and limited interactions with the tracker. In our model we consider different factors that influence the quality of user experience and train the proposed model on diverse state-action transitions when viewers interact with the tracker. In addition, provided that past events have various user experience characteristics we follow a gradient boosting strategy to compute a global model that learns from different events. Our experiments with three real-world datasets of live video streaming events demonstrate the superiority of the proposed model against several baseline strategies. Moreover, as the majority of the viewers at the beginning of an event has poor experience, we show that our model can significantly increase the number of viewers with high quality experience by at least 75\% over the first streaming minutes. Our evaluation datasets and implementation are publicly available at \url{https://publicresearch.z13.web.core.windows.net}
\end{abstract}

%



\maketitle
\pagestyle{plain}

\section{Introduction} \label{sec:intro}


Live video streaming has been established as a primary communication service during the digital transformation of enterprises. As part of the internal communication among employees, Fortune-500 companies on a daily basis schedule live video streaming events such as virtual town halls, employees' education through webinars, and so on. Accounting for the high value of live video streaming services, companies invest a significant amount of their annual budgets so that every employee who attends an event can have a similar high-quality user experience as in person communication~\cite{ibmadoptionreport}. However, in the real-world setting there are several factors hindering a high-quality user experience, for example, the bandwidth limitations of several offices when several employees attend an event simultaneously. This results in a low user experience which not only negatively impacts the employee's engagement, but also the event's return of investment \cite{ibmreport, hiveengagement, dobrian2011understanding}. 

\begin{figure}
    \centering
    \includegraphics[scale=0.35]{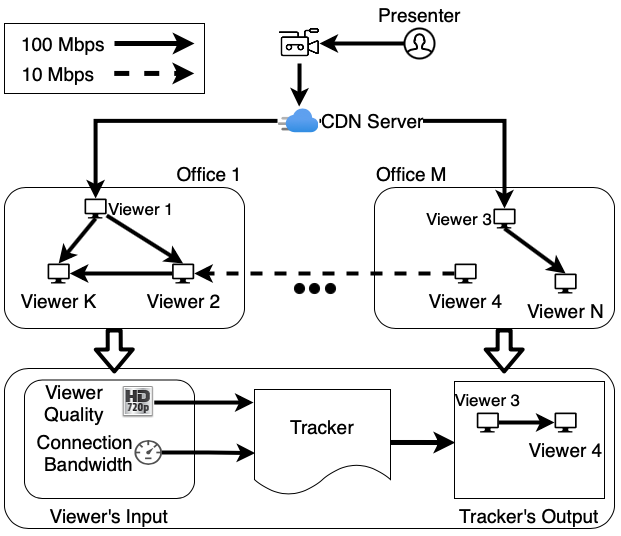}
    \caption{In a live video streaming event, a viewer periodically reports the connection bandwidth as well as her quality of experience to the tracker which then probes the viewers to adjust their connections accordingly.}
    \label{fig:distribution}
\end{figure}

In an attempt to address this problem large organizations focus on distributed live video streaming solutions, trying to overcome the offices' bandwidth limitation by reducing the number of viewers in each office that simultaneously fetch the video stream directly by the Content Delivery Network (CDN) server~\cite{Roverso2015, zhao2005gridmedia}. Having established a limited number of connections, when a viewer receives the video stream she distributes it to the rest of the viewers. Distributed solutions aim at exploiting the internal high-bandwidth network to improve the user experience. As shown in Figure \ref{fig:distribution}, viewers $1$ and $3$ are connected directly to the CDN server to download the video stream in the offices $1$ and $M$, respectively. Thereafter, viewer $1$ distributes the video content to viewers $2$ and $K$, and viewer $3$ to viewer $N$ via high-bandwidth connections ($100$ Mbps). In practice though, immediately selecting the high-bandwidth connections among the viewers is not always feasible. The main reason is the absence of prior knowledge of the organization's large-scale enterprise network, for instance the information that viewers $3$, $4$ and $N$ are located in the same office, due to privacy issues and network's upgrading tasks \cite{antaris2020vstreamdrls}. Given the information of a limited number of connections, distributed solutions have to predict the high-bandwidth connections in real-time during the live video streaming event. In live video streaming events, distributed solutions set up a centralized server - tracker to select the viewer's connections. The goal of the tracker is to detect the proper connections, allowing viewers to have a high-quality user experience. To achieve this, the tracker monitors various data that are periodically reported by viewers, such as connection bandwidth, viewer's player quality, and so on. Then, the tracker processes these data to predict the network capacity among viewers and select the high-bandwidth connections. For example, given the predicted network capacity among viewers in Figure \ref{fig:distribution}, the tracker adjusts the connections of viewer $4$ and probes viewer $3$ to link to viewer $4$ with a high-bandwidth connection. 

When an event starts, several viewers join at the same time. Selecting the connections that will allow each viewer to have a high-quality user experience from the first minutes of the event is a challenging task, as most of the viewers initially establish low-bandwidth connections. This happens because the tracker requires a large amount of viewers' interactions, to accurately predict the high-bandwidth connections among the viewers. Therefore, the majority of the viewers have a poor user experience at the beginning of the event. In addition, the bandwidth capacity among viewers might change over time, as viewers emerge at unfavorable locations with low-bandwidth connections during the event \cite{antaris2020vstreamdrls, Ahmed2017qoe}. As a consequence, viewers tend to leave the event when their user experience is significantly reduced \cite{Wassermann2019engagement, Ahmed2017qoe, ibmadoptionreport}. Therefore, it is essential not only to predict the user experience in real-time, but also to predict the proper connections to improve the user experience.

In a live video streaming event, the viewers' network can be modeled as an evolving graph, where the nodes/viewers adjust their edges/connections based on the tracker selection at each minute of the event. In addition, viewers are tagged with dynamic labels such as bad, poor, fair, good, and excellent, corresponding to the user experience per viewer at a certain minute of the event  \cite{Laghari2012qoe, yin2015qoe, Zhang2020}. The goal of the tracker is to predict the viewers' dynamic labels and select their connections accordingly aiming to improve the user experience. In the node classification task, evolving graph representation learning strategies compute low-dimensional node embeddings to classify the nodes' dynamic labels. Baseline approaches run random walks on consecutive graph snapshots to capture the evolution of the graph \cite{nguyen2018ctdne}, other approaches apply self-attention mechanisms \cite{sankar2020dysat, antaris2020vstreamdrls, xu2020tgat}, recurrent neural networks (RNNs) \cite{kumar2019jodie, pareja2020evolvegcn}, and Long-Short Term Memory (LSTM) units \cite{Wang2020dygnn}. However, as we will demonstrate in Section \ref{sec:experiments} baseline evolving graph representation learning strategies require a significant amount of viewers' connections to achieve high prediction accuracy in the user experience. This occurs because at the beginning of the event most viewers are connected with low-bandwidth connections, thus baseline strategies mainly focus on accurately predicting the poor user experience. Moreover, such approaches are not designed to select the connections that will improve the user experience in the next minute of the event. As a consequence they do not necessarily work on streaming events, where given a limited number of viewers' interactions at the beginning of an event the tracker needs to predict the viewers' experience at the next minute and exploit the predictions to adjust the connections accordingly. In addition, learning from various past events can also leverage the prediction of user experience from evolving networks with different sets of users. In several applications boosting strategies have proven an effective means for improving the accuracy of weak classifiers~\cite{MasonBBF99,Schapire90}. Nonetheless, the influence of boosting strategies on user experience based on past streaming events has not been studied yet. 

To face the user experience challenges, we propose a \textbf{G}raph r\textbf{E}inforcement \textbf{L}earning model for live video \textbf{S}treaming events (GELS), making the following contributions: 

\begin{itemize}
\item We address the user experience prediction task as a classification problem and solve it with deep reinforcement learning, formulating an event as a Markov Decision Process (MDP) and accounting for different factors that influence the user experience in live video streaming events. 
\item We train our model on the most diverse viewers' transitions in user experience, and then exploit the predicted labels to select the connections that will improve the user experience at the first minutes of the event. 
\item We present a gradient boosting strategy to extract information from past events and improve the user experience via GELS over several video streams.
\end{itemize}

The remainder of this paper is organized as follows, in Section \ref{sec:problem} we detail the proposed GELS model and in Section \ref{sec:data} we analyse the evolution of user experience in real-world events. Then, in Section \ref{sec:experiments} we present the experimental evaluation of our model against baseline strategies, and conclude the study in Section \ref{sec:conclusions}.

\section{Proposed Model} \label{sec:problem}

\subsection{Quality of User Experience in Live Video Streaming Events}

A streaming event is modeled as an evolving graph $\mathcal{G} = (\mathcal{V}, \mathcal{E}^t)$. At each minute $t = 1, \ldots, T$ of the event, $n$ viewers of set $\mathcal{V} = \{u_1, u_2, \ldots v_n\}$ interact with the tracker to select connections $e^{t+1}_{u,v} \in \mathcal{E}^{t+1}$ between viewers $u$ and $v \in \mathcal{V}$ in the next minute $t+1$. We formulate the selection process of the viewers' connections during a live video streaming event as a Markov Decision Process (MDP), which is defined as $M = (\mathcal{S}, \mathcal{A}, \mathcal{P}, \mathcal{R}, \gamma)$, with $\mathcal{S}$, $\mathcal{A}$, $\mathcal{P}$ being the state, action and transition probability sets, and $\gamma$ the discount factor. At the $t$-th minute, the tracker/agent takes an action $a^t \in \mathcal{A}$ to connect viewers $u$ and $v$ based on a state $s^t_u \in \mathcal{S}$ of viewer $u$. At the next minute $t+1$, the tracker receives a reward $R(s^{t},a^{t}) \in \mathcal{R}$ for the previously selected action $a^t$ and the state $s^t_u$ of viewer $u$ is updated to $s^{t+1}$ with transition probability $p(s^{t+1}_u|s^t_u,a^t) \in  \mathcal{P}$. The reward $R(s^{t},a^{t})$ corresponds to the user experience $QoE^{t}_{u,v}$ of viewer $u$, given her connection $e^t_{u,v}$ with viewer $v$. 

In the reinforcement learning setting, we consider the user experience $QoE^t_{u,v}$ as the reward that viewer $u$ provides to the tracker. We first divide the live video stream in multiple segments $c$ \cite{Lucian2010http,Gao2018qoe}, and then calculate the user experience $QoE^t_{u,v}$ of the connection $e^t_{u,v}$ at the $t$-th minute by considering the following factors \cite{yin2015qoe, Mao2017pensieve}:

\begin{itemize} 
    \item \textit{Average Video Quality}: For each viewer $u$ the average quality $q$ of the $K$ video segments distributed by viewer $v$:  \begin{equation}\frac{1}{K} \sum_{k=1}^{K} q(c_k) \end{equation}
    \item \textit{Average Quality Variations}: The magnitude of changes in the video quality between consecutive segments: \begin{equation}\frac{1}{K-1} \sum_{k=1}^{K-1} |q(c_{k+1}) - q(c_k)|\end{equation}
    \item \textit{Rebuffer}: The number of player rebufferings occurred to viewer $u$ when receiving the video segment $c$. Given the size $d_k(c_k)$ of a segment $c$ and the bandwidth $b^t_u$ of viewer $u$ at the $t$-th minute, a rebuffering occurs when the download time $\frac{d_k(c_k)}{b^t_u}$ is larger than the player buffer size $B$: 
    \begin{equation}\sum_{k=1}^{K} 1 \bigg(\frac{d_k(c_k)}{b^t_u} > B^t \bigg)\end{equation}
\end{itemize}
Accounting for the quality factors in Equations 1-3, we define the user experience $QoE^t_{u,v} \in \mathbb{R}^{+}$ of connection $e^t_{u,v}$ between viewers $u$ and $v$ at the $t$-th minute as follows:

\begin{equation}
\begin{split}
QoE^t_{u,v} & = \frac{1}{K} \sum_{k=1}^{K} q(c_k) - \lambda \frac{1}{K-1} \sum_{k=1}^{K-1} |q(c_{k+1}) - q(c_k)| \\
 &  - \mu \sum_{k=1}^{K} 1 \bigg(\frac{d_k(c_k)}{b^t_u} > B^t \bigg)
\end{split}
\label{eq:qoe}
\end{equation}
where $\lambda$ and $\mu$ are non-negative weighting parameters to balance the importance of each factor in the user experience. A low $\lambda$ value indicates that the user experience is more affected by the average video quality than the variability of the quality among consecutive segments. For large $\mu$ values the user experience emphasizes more on the rebuffering occurrences during the streaming event. Given the user experience $QoE^t_{u,v}$ between  viewers $u$ and $v$ at the $t$-th minute, the objective of the tracker/agent is to compute an optimal policy $\pi_{\theta}: \mathcal{S}\times\mathcal{A} \rightarrow [0,1]$, where $\theta$ are the parameters of policy $\pi$. To learn the optimal policy $\pi_{\theta}$, the tracker optimizes the Bellman equation, by maximizing the state-action transition value $Q_w(s^t,a^t) = \max_{\pi_{\theta}} \mathbb{E}_{\pi_{\theta}}\{\sum_{x=0}^{\infty}\gamma^{x}R(s^t, a^t)\}$ \cite{bellman1966dynamic}. The state-action transition value $Q_w(s^t,a^t)$, parameterized by $w$, represents the expected value of the action $a^t$  based on the policy $\pi_{\theta}$.

\subsection{Improving User Experience via GELS}

\begin{figure}
    \centering
    \includegraphics[scale=0.35]{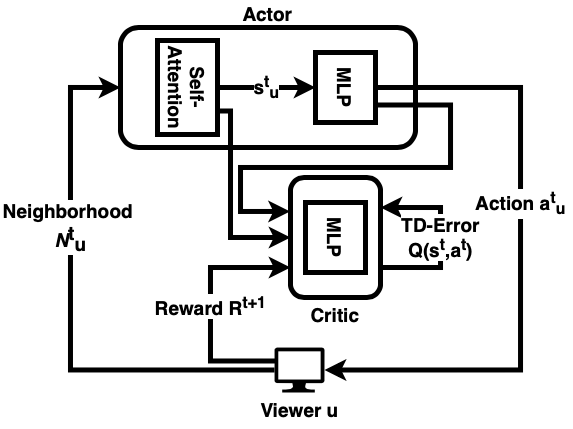}
    \caption{Overview of the viewer's interaction with the tracker via GELS at the $t$-th minute.}
    \label{fig:gels}
\end{figure}

To find the optimal policy $\pi_{\theta}$, we employ the actor-critic scheme of reinforcement learning \cite{konda2000actor, sutton2018rl}. In Figure \ref{fig:gels}, we illustrate an overview of the interaction of viewer $u$ with the tracker via GELS. At the $t$-th minute of the event, viewer $u$ reports the neighborhood $\mathcal{N}^t_u$ to the tracker. The actor network computes the state vector $\mathbf{s}^t_u \in \mathbb{R}^d$ and outputs the action vector $\mathbf{a}^t_u \in \mathbb{R}^d$. The tracker adjusts the connections of viewer $u$, by selecting the action $a^t_u$ with the highest value in the action vector $\mathbf{a}^t_u $. Then, viewer $u$ adjusts her connections and gives a reward $R(s^t, a^t)$ of the action $a^t$ for the next minute $t+1$. The critic network takes as input the state vector $\mathbf{s}^t$ and action vector $\mathbf{a}^t$ to estimate the state-action value $Q(s^t, a^t)$. The critic network exploits the estimated state-action value $Q(s^t, a^t)$ and the reward $R(s^t, a^t)$, to optimize the parameters of the model based on a temporal-difference error $L$ (Equation \ref{eq:loss}). In doing so, the tracker evaluates: i) the performance of the critic network to predict the user experience of viewer $u$ at the next minute $t+1$, and ii) the ability of the actor network to compute the actions at the $t$-th minute to improve  the experience of viewer $u$ at the next minute $t+1$.

\subsection{Deep Graph Reinforcement Learning} \label{sec:dgrl}

In the actor-critic model, at each minute $t = 1,\ldots, T$ the input of the actor model is the neighborhood $\mathcal{N}^t_u$ of viewer $u$. Provided that viewers have no features, we compute an embedding lookup to extract the node embedding $\mathbf{z}_u \mathbb{R}^l$ for each viewer $u $ \cite{Lee2019, Lu2020metahin}. Given the node embeddings $\mathbf{z}_v$ of each neighbor-viewer $v \in \mathcal{N}^t_u$, the actor network computes a state vector $\mathbf{s}^t_u$, which corresponds to the state $s^t_u$ of viewer $u$ at the $t$-th minute. We calculate the state vector $\mathbf{s}^t_u$ based on a self-attention mechanism as follows~\cite{Vaswani2017attention}: $\mathbf{s}^t_u = \sigma\bigg(\sum_{v\in \mathcal{N}_u} o_{u,v} \mathbf{W} \mathbf{z}_u\bigg)$, where $\sigma(\cdot)$ is the non-linear activation function ReLU, and $\mathbf{W} \in \mathbb{R}^{d \times l}$ is the weight parameter matrix. Variable $o_{u,v}$ corresponds to the normalized attention coefficient between $u \in \mathcal{V}$ and $v \in \mathcal{N}^t_u$, which is calculated according to the softmax function as follows: $$o_{u,v} = \frac{exp\bigg(\sigma(QoE^t_{u,v} \cdot \alpha^{\top} [\mathbf{W} \mathbf{z}_u || \mathbf{W} \mathbf{z}_v])\bigg)}{\sum_{y \in \mathcal{N}_u} exp\bigg(\sigma(QoE^t_{u,y} \cdot \alpha^{\top} [\mathbf{W} \mathbf{z}_u || \mathbf{W} \mathbf{z}_y])\bigg)}$$ where $\alpha^{\top}$ is the $2d$-dimensional weight vector applied to the attention process. The symbol $||$ denotes the concatenation among the transformed node embeddings $\mathbf{z}_u$ and $\mathbf{z}_v$. Given the state $\mathbf{s}^t_u$ of $u$, we have to compute the action vector $\mathbf{a}_u$, which reflects on the probability of improving the experience of viewer $u$ when connecting with viewer $v$. For each viewer $u$, we consider all the viewers $\mathcal{V} \setminus \{u\}$ as the set of possible actions $\mathcal{A}$ at the $t$-th minute. The action vector $\mathbf{a}^{t}_u$ is computed  by employing a two-layer perceptron as follows: $\mathbf{a}^t_u = \sigma\bigg(\mathbf{W}^2 ReLU(\mathbf{W}^1 \mathbf{s}^t_u)\bigg)$, with $\mathbf{W}^1 \in \mathbb{R}^{m \times d}$ and $\mathbf{W}^2 \in \mathbb{R}^{n \times m}$ being the weight matrices of the first and second layers, and $m$ the intermediate dimensions between the two layers. In our implementation, we select the optimal action, that is the action of connecting $u$ with the viewer with the highest probability of improving the user experience, by employing the softmax function on the action vector $\mathbf{a}^t_u$. 

To evaluate the performance of the actor network when selecting the respective actions for improving the user experience, we adopt the critic network. Having computed the action $a^t_u$ and the state $s^t_u$ at the $t$-th minute, the role of the critic network is to estimate the improvement of the experience of viewer $u$ at the next minute $t+1$. Therefore, the input of the critic network is the state vector $\mathbf{s}^t_u$ and the action vector $\mathbf{a}^t_u$ of the node $u$ at the $t$-th minute. The critic network outputs a scalar state-action value $Q_w(\mathbf{s}^t_u, \mathbf{a}^t_u)$, parameterized by $w$, which is an approximation of the true state-action value function, namely the Q-value function. We calculate the state-action value by employing a two-layer perceptron as follows: $Q_w(\mathbf{s}^t_u, \mathbf{a}^t_u) = MLP(\mathbf{s}^t_u, \mathbf{a}^t_u)$. 

The actor-critic network is optimized based on a replay memory buffer, which contains the $k$ latest state-action transitions, resulting in the most diverse user experience. This is achieved by exploiting the KL-divergence on the $QoE^t_u$ value of each viewer $u\in\mathcal{V}$ between consecutive minutes as follows \cite{Kullback2015}: 
\begin{equation} \label{eq:kldiv}
    KL(QoE^t_{u}|QoE^{t+1}_u) = \sum_{v\in \mathcal{N}_u} QoE^{t+1}_{u,v} log \bigg(\frac{QoE^{t+1}_{u,v}}{QoE^{t}_{u,v}}\bigg)
\end{equation}
This means that the replay memory buffer contains the state-action transitions of the viewers which significantly change the user experience between consecutive minutes. As we will demonstrate in Section \ref{sec:experiments}, in doing so the tracker/agent learns the optimal policy $\pi_{\theta}$ at the first minutes of the event, allowing the viewers to achieve a high-quality user experience. Based on the replay memory buffer, we optimize the parameters of the actor and critic model, following the temporal-difference learning strategy \cite{sutton2018rl}, by minimizing the following mean-squared error loss function:

\begin{equation}
    L = \frac{1}{2} \sum^T_{t=1} \mathbb{E}_{\mathbf{s}^t,\mathbf{a}^t} [(\mathbb{E}_{\mathbf{s}^{t+1}, R}[R(\mathbf{s}^t,\mathbf{a}^t) + \gamma Q(\mathbf{s}^{t+1},\mathbf{a}^t)] - Q(\mathbf{s}^t,\mathbf{a}^t))^2]
    \label{eq:loss}
\end{equation}

\subsection{Gradient Boosting Strategy} \label{sec:grad}

To improve the user experience via GELS over several live video streaming events and learn a global policy $\pi_{\theta}$, we follow a gradient boosting strategy \cite{Schapire1990boost, Mason1999boost}. In particular, we exploit a set $\mathcal{M}$ of historical events and formulate each event as a MDP $M_i \in \mathcal{M}$. For each event $M_i$, we divide the connections $\mathcal{E}_i$ in two sets, the training set $\mathcal{E}_i^{tr}$ and the test set $\mathcal{E}_i^{ts}$. The training set $\mathcal{E}_i^{tr}$ is used to adjust the policy $\pi_{\theta}$ to an event $M_i$. This is achieved by adopting the actor-critic network of Section~\ref{sec:dgrl} to learn the optimal policy $\pi_{\theta}$ on each event. The actor network selects actions based on the connections that exist in the training set $\mathcal{E}_i^{tr}$. To derive a global model over several events, we adjust the parameters $\phi$ and $w$ of the actor and critic network, respectively, by adopting a boosting strategy on the test set $\mathcal{E}_i^{ts}$. More specifically, we employ back-propagation on the parameters $\phi$ and $w$ as follows:

\begin{equation}
\begin{array}{c}
      \phi \leftarrow \phi - \eta \frac{\partial L_{\mathcal{E}_i^{ts}}(\mathbf{a}^t, Q_w(\mathbf{s}^t, \mathbf{a}^t), \mathcal{E}_i^{ts})}{\partial \phi}  \\ \\
      w \leftarrow w - \eta \frac{\partial L_{\mathcal{E}_i^{ts}}(\mathbf{a}^t, Q_w(\mathbf{s}^t, \mathbf{a}^t), \mathcal{E}_i^{ts})}{\partial w}
\end{array}
\end{equation}
where $L$ is the loss function in Equation \ref{eq:loss}. Finally, we generate a global model over the set  $\mathcal{M}$ of different events, by minimizing the loss function over all the events $M_i \in \mathcal{M}$ as follows:

\begin{equation}
    \min_{\theta} \sum_{M_i \in \mathcal{M}} L_{E_i^{ts}}(a, Q, E_i^{ts})
\end{equation}

\section{User Experience in Live Video Streaming Events} \label{sec:data}
\textbf{Evaluation Datasets}
We collected three real-world datasets, that is the LiveStream-$1$, LiveStream-$2$ and LiveStream-$3$ datasets, which were anonymized and made publicly available. Each dataset consists of $30$ evolving graphs, corresponding to $30$ different live video streaming events occurred in large enterprise networks. The duration of each event was $45$ minutes. To generate each evolving graph, we monitored the viewers' connections and their interactions with the tracker during the events. The LiveStream-$1$ dataset consists of $24,752$ viewers and $1.8$ million connections over $30$ events. Each viewer participated in $2.7$ different events on average, and the viewers were distributed to $62$ offices of the same enterprise network. The LiveStream-$2$ dataset consists of $28,879$ viewers in total, who were spread to $12$ offices of the same organization. Each viewer participated in $1.3$ events on average, and the viewers established $590,000$ connections in total. The LiveStream-$3$ dataset was generated based on $30$ events of $15$ different enterprise networks. The dataset consists of $148,982$ viewers and $4.1$ million connections. Each viewer participated in $1.1$ events on average and the viewers were distributed to $56$ offices.

\begin{figure*}[ht]
    \centering
    \includegraphics[scale=0.24]{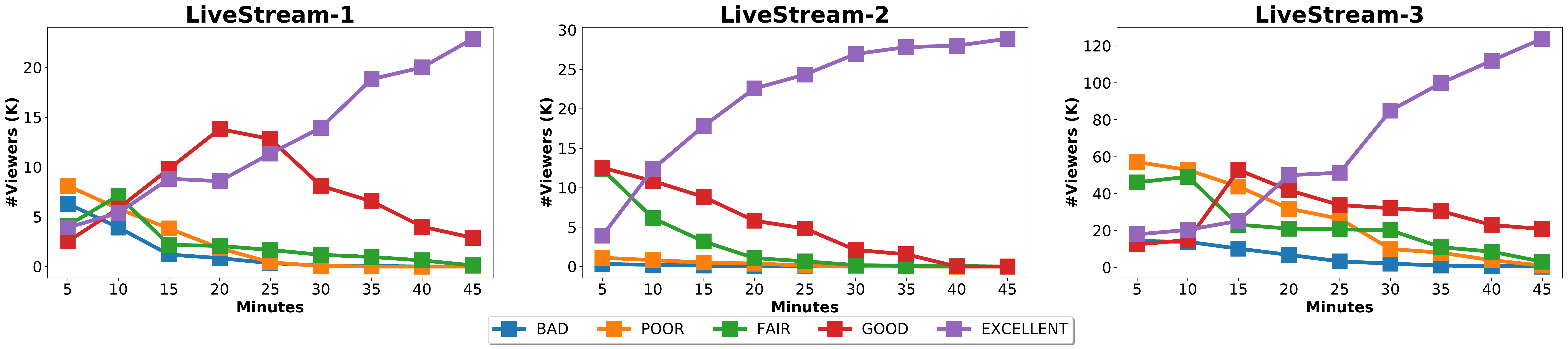}
    \vspace{-0.05cm}
    \caption{Average \#viewers of each user experience label during the events.}
    \label{fig:qoe_viewers}
\end{figure*}

\begin{figure*}[ht]
    \centering
    \includegraphics[scale=0.24]{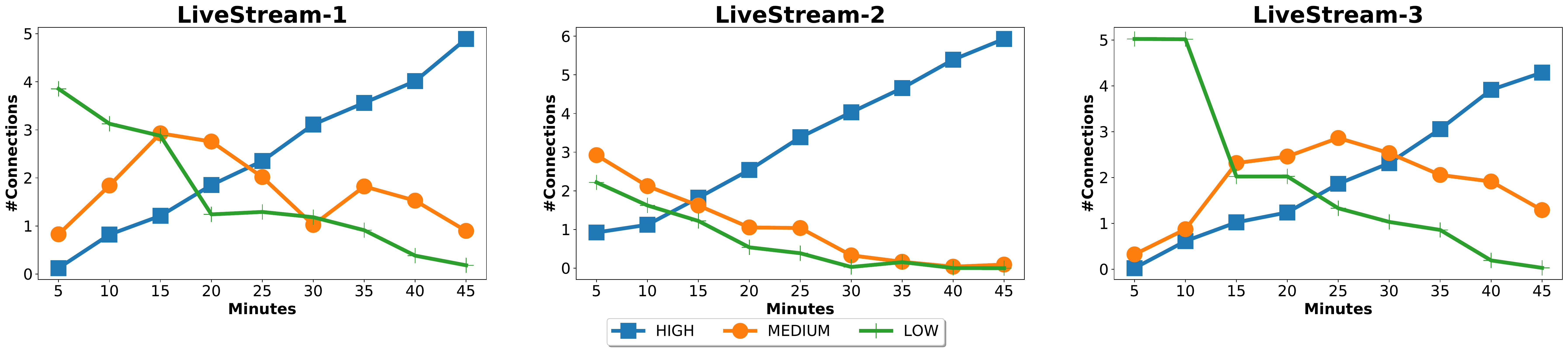}
    \vspace{-0.05cm}
    \caption{Average \#high-bandwidth connections established during the live video streaming events.}
    \label{fig:connections}
\end{figure*}

\begin{figure*}
    \centering
    \begin{tabular}{ccc}
        \includegraphics[scale=0.21]{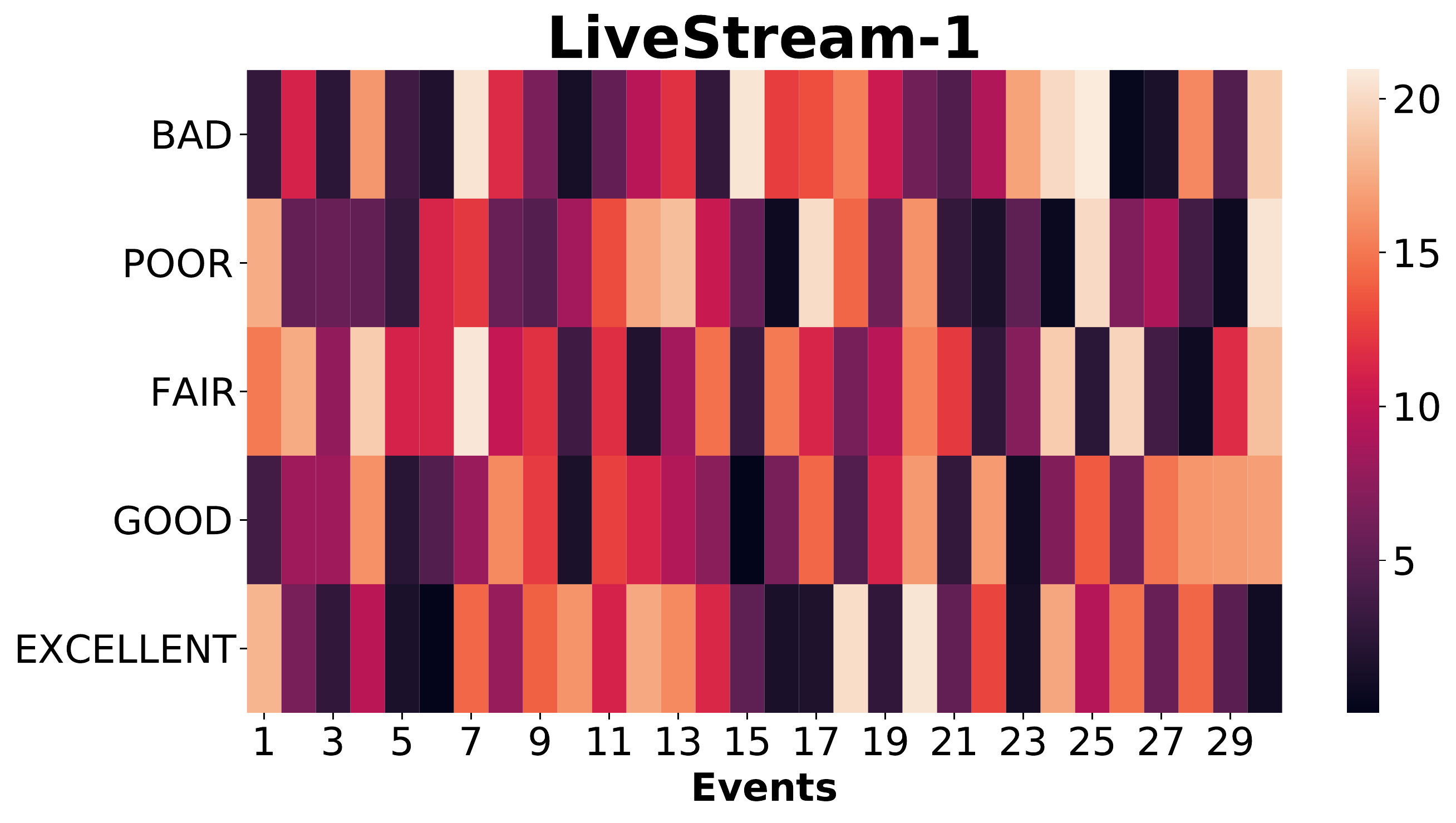} &  \includegraphics[scale=0.21]{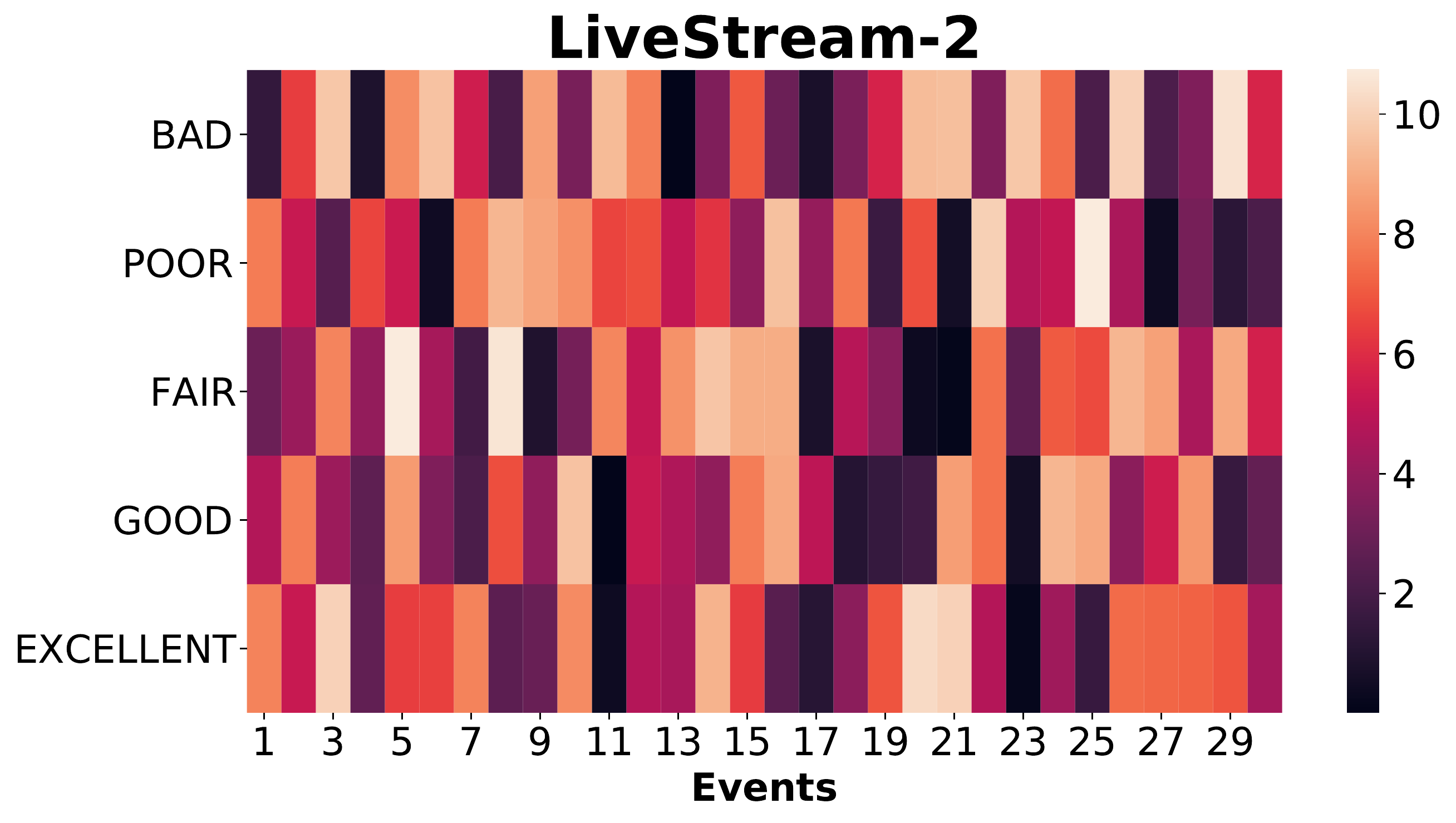} & \includegraphics[scale=0.21]{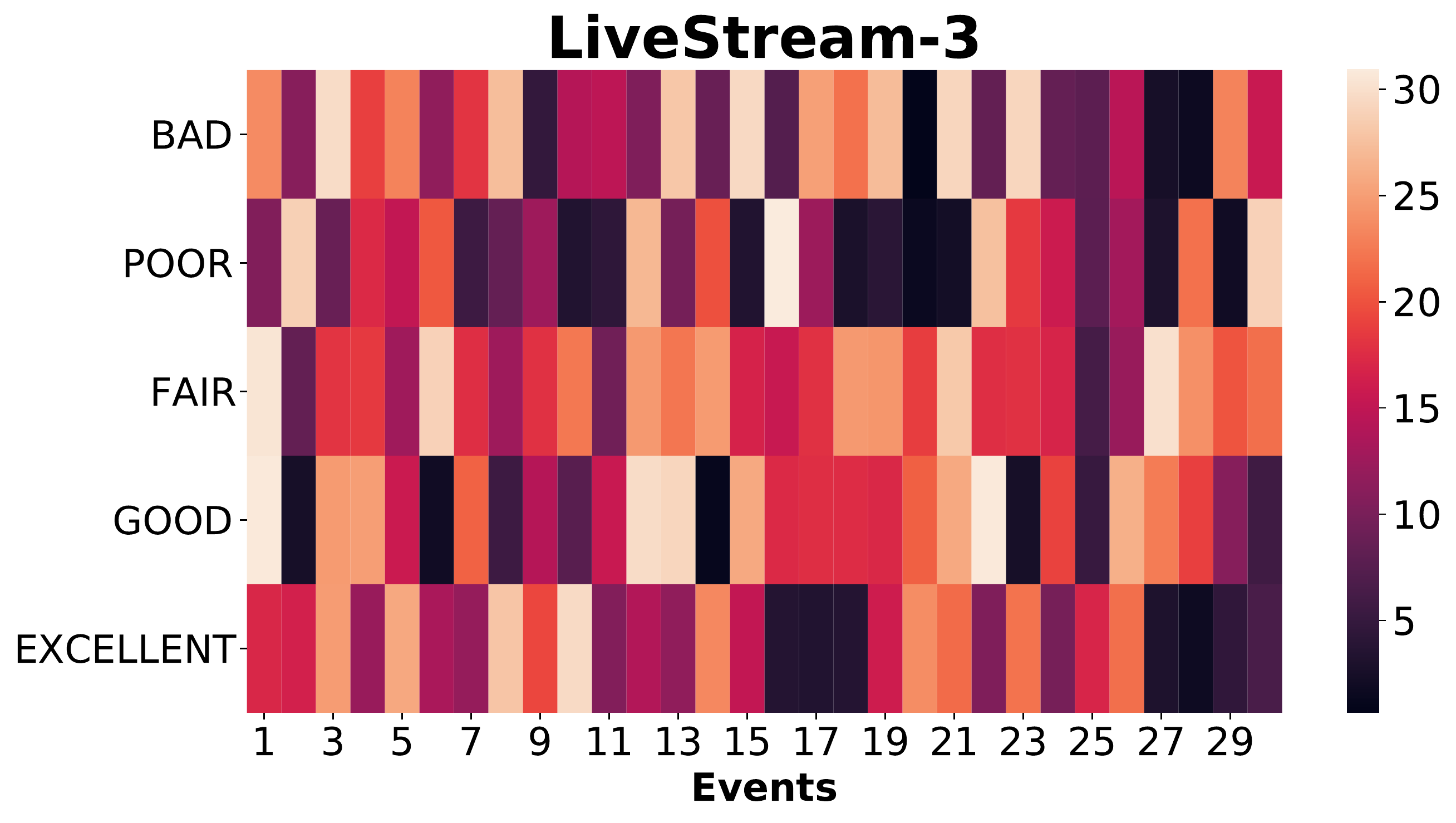}
    \end{tabular}
    \vspace{-0.05cm}
    \caption{Average distribution of KL-divergence of the user experience labels during the events.}
    \label{fig:qoe_kl}
\end{figure*}

\textbf{Evolution of User Experience.}
To generate the different user experience labels, we normalized the $QoE^t_u$ value of each viewer $u$ based on the maximum possible $QoE$ value that the viewer $u$ could have achieved at the $t$-th minute, that is when the viewer $u$ attended at the highest video quality without having quality variations and rebufferings. Given the normalized user experience, we performed binning of equal-length in the $QoE$ values and assigned the following user experience labels to the viewers: bad, poor, fair, good and excellent for every streaming minute \cite{Laghari2012qoe, yin2015qoe, Zhang2020}. In Figure \ref{fig:qoe_viewers}, we present the average number of viewers for each user experience label over the streaming minutes. We observe that in the LiveStream-$1$ and LiveStream-$3$ datasets, most of the viewers achieved good and excellent user experience after the first $10$ and $15$ streaming minutes, respectively. Moreover, the tracker required more than $20$ streaming minutes to select the optimal viewers' connections and achieve excellent user experience for the majority of the viewers. This indicates that the tracker required a large amount of interactions with the viewers to identify the high-bandwidth connections. In Livestream-$2$, the majority of the viewers had fair or good user experience at the first $5$ streaming minutes. Most of the viewers achieved excellent user experience after $10$ streaming minutes. This means that the majority of the connections in LiveStream-$2$ exhibited high network capacity, allowing the viewers to efficiently distribute the video content and achieve excellent user experience.

\textbf{Viewers' Connection Level.}
As aforementioned in Section \ref{sec:intro}, during a live video streaming event each viewer has a limited number of connections. Therefore, it is essential for the tracker to identify the high-bandwidth connections at the first minutes of the event and achieve excellent user experience. In the evaluation datasets we categorized each connection between two viewers as low, medium and high based on the connection throughput. We applied binning on the connection throughput as follows: $<5$ Mbps throughput were considered as low connections, $5-10$ Mbps medium, and  $>10$ Mbps  high. As illustrated in Figure \ref{fig:connections}, in LiveStream-$1$ and LiveStream-$3$ datasets, most of the viewers had less than $1$ high-bandwidth connections in the first $10$ minutes. After the first $10$ minutes, the viewers managed to establish more than $1$ high-bandwidth connections. Instead, in LiveStream-2 most of the viewers were able to connect to more than $1$ high-bandwidth connections at the first $5$ minutes, indicating that the majority of the viewers in LiveStream-2 were located either in the same office, such as the headquarters of the organization, or in different offices connected with a high-bandwidth network such as a fiber optic network \cite{wong2020telecommunications}.

\textbf{KL-Divergence Distribution.}
In Figure \ref{fig:qoe_kl}, we present the average distribution of the KL- divergence of the user experience during the events (Equation~\ref{eq:kldiv}). We observe that in all datasets the viewers had significant differences in the user experience, as each live video streaming event occurred in different offices of the organization with various network capacity characteristics. Therefore, the tracker had to adjust the connections' selection policy in real-time so as to achieve excellent user experience. Moreover, we observe that in all datasets, most of the events have high KL-divergence in the good and excellent user experience labels. This means that the viewers' experiences were significantly changing  during the streaming minutes.

\section{Experiments} \label{sec:experiments}

\subsection{Setup}
\textbf{Evaluation Protocol}
In our experiments, we evaluate the performance of the proposed GELS model on the user experience classification task. Given the viewers' connections and user experience labels at the $t$-th minute, the goal of the classification task is to predict the user experience label that each viewer will have in the next minute $t+1$. Provided that we have multiple user experience labels (Section~\ref{sec:data}), we follow a One-vs-Rest strategy to perform binary classification for each user experience label. The classification accuracy of our model is evaluated based on the area under the ROC curve (AUC), Micro-F1 and Macro-F1 metrics \cite{fmeasure, FAWCETT2006861}. AUC represents the degree of separability of each user experience label, Micro-F1 emphasizes on the common labels, while Macro-F1 penalizes models with low accuracy in the rare labels. Following the evaluation protocol of \cite{pareja2020evolvegcn, sankar2020dysat, Lu2020metahin}, the connections and user experience labels of each event are divided into the training  $\mathcal{E}^{tr}$ and test sets $\mathcal{E}^{ts}$. The training set $\mathcal{E}^{tr}$ consists of all  the user experience labels and connections that viewers established in the first $30$ streaming minutes and the rest of user experience labels and connections correspond to the test set $\mathcal{E}^{ts}$. In each dataset, we randomly select $26$ events to train our model, $2$ events are considered for validation and the remaining $2$ events for testing. The goal of the classification task of user experience is to predict the label of the viewers in the test events. We repeated our experiments five times and report average AUC, Micro-F1 and Macro-F1.  

\textbf{Baseline Strategies} We evaluate the performance of the proposed GELS model against the following strategies: \textbf{DyGNN}\footnote{\url{https://github.com/alge24/dygnn}} \cite{Ma2020dygnn} a graph neural network that captures the evolution of the graph by adopting LSTM units in the evolving connections. \textbf{EvolveGCN}\footnote{\url{https://github.com/IBM/EvolveGCN}} \cite{pareja2020evolvegcn} a dynamic learning approach that adopts Gated Recurrent Units (GRUs) on the weights of consecutive Graph Convolutional Networks (GCNs). \textbf{DySAT}\footnote{\url{https://github.com/aravindsankar28/DySAT}} \cite{sankar2020dysat} a graph representation learning strategy that employs a multi-head self-attention mechanism to compute the graph's structural and temporal evolution. \textbf{VStreamDRLS}\footnote{\url{https://github.com/stefanosantaris/vstreamdrls}} \cite{antaris2020vstreamdrls} a deep neural network model that applies a self-attention mechanism to calculate the evolution of the graph on the weights of consecutive GCN models. \textbf{MetaHIN}\footnote{\url{https://github.com/rootlu/MetaHIN}} \cite{Lu2020metahin} a meta-learning strategy on evolving graphs that captures the different semantic facets of each node/viewer. \textbf{PolicyGNN}\footnote{\url{https://github.com/lhenry15/Policy-GNN}} \cite{lai2020policygnn} a graph neural network strategy that performs reinforcement learning to compute the required number of layers in the GCNs. \textbf{GELS*}\footnote{\url{https://publicresearch.z13.web.core.windows.net}} a variant of our model, which ignores the gradient boosting strategy of past events in Section~\ref{sec:grad} and trains the actor-critic model on a single event.

\textbf{Environment.} All our experiments were conducted on a single server with an Intel Xeon Bronze 3106, 1.70GHz CPU. We accelerated the modes' training using the GPU Geforce RTX 2080 Ti graphic card. The operating system of the server was Ubuntu 18.04 LTS. We implemented our proposed GELS model in Pytorch 1.7.1 and created the reinforcement learning environment using the OpenAI Gym 0.17.3 library. To measure the user experience, we emulated the video player of each viewer based on the dash.js library \cite{dashjs}. For the models' training on each event, the reinforcement learning environment was initialized using the training set $\mathcal{E}^{tr}$. This means that the tracker selected an action $a^t$ to establish a connection between viewers $u$ and $v $ based on the training set $\mathcal{E}^{tr}$ at each minute $t$ of the event. Each viewer monitors (i) the quality of the video, reported by the dash.js library, (ii) the connection throughput and (iii) the reward returned to the agent, that is the $QoE^t_{u,v}$ value of the connection $e_{u,v} \in \mathcal{E}^{tr}$. Similarly,  the reinforcement learning environment was initialized to evaluate the performance of the learned policy $\pi_{\theta}$ in the test set $\mathcal{E}^{ts}$.

\textbf{Configuration.} For each examined model, we tuned the hyper-parameters following a grid-selection strategy on the validation set, and report the performance based on the best configuration. In DyGNN we set the node embedding dimension to $256$ in all datasets. In EvolveGCN and DySAT the node embedding is fixed to $128$ and $256$ in all datasets, respectively. Moreover, EvolveGCN exploits the information from $2$ consecutive graph snapshots, while DySAT from $3$ snapshots in all datasets. In MetaHIN, we use $128$-dimensional node embeddings for LiveStream-$1$ and LiveStream-$3$, and $32$-dimensional embeddings for Livestream-$2$. PolicyGNN uses $64$-dimensional embeddings in all datasets, the discount factor $\gamma$ is set to $0.92$ and the $\epsilon$ value in the $\epsilon$-greed exploration-exploitation strategy is fixed to $0.1$. In the proposed GELS model and its variant GELS*, we fix the node embeddings size to $128$ in all datasets. The size of the replay memory buffer is set to $k$=$64$ state-action transitions, the discount factor $\gamma$ is $0.96$ and the $\epsilon$ value is fixed to $0.2$. To calculate the $QoE$ value in Equation \ref{eq:qoe}, we fix the user experience parameters to $\lambda$=$0.2$ and $\mu$=$0.3$ \cite{yin2015qoe}.



\subsection{Performance Comparison over Streaming}

\begin{table*}[t]
    \caption{Methods' comparison in terms of AUC, Micro-F1 and Macro-F1 over the streaming minutes. Bold values indicate the best method using a statistical significance t-test with $p<0.01$.}
    \centering
    \resizebox{\textwidth}{10cm}{
    \begin{tabular}{c|c|c|c|c|c|c|c|c|c|c}
         & \textbf{Baselines} & \multicolumn{9}{c}{\textbf{Streaming Minutes}}  \\ \hline
        \textbf{LiveStream-1} & & $5$ & $10$ & $15$ & $20$ & $25$ & $30$ & $35$ & $40$ & $45$ \\ \hline
        \multirow{8}{*}{\textbf{AUC}}       & \textbf{DyGNN}        & $.492 \pm .355$ & $.589 \pm .382$ & $.663 \pm .285$ & $.776 \pm .241$ & $.827 \pm .216$ & $.844 \pm .316$ & $.843 \pm .336$ & $.849 \pm .371$ & $.845 \pm .319$ \\
                                            &  \textbf{EvolveGCN}   & $.513 \pm .112$ & $.597 \pm .162$ & $.614 \pm .171$ & $.637 \pm .188$ & $.661 \pm .192$ & $.682 \pm .201$ & $.715 \pm .105$ & $.712 \pm .150$ & $.719 \pm .182$ \\
                                            &  \textbf{DySAT}       & $.614 \pm .231$ & $.693 \pm .245$ & $.714 \pm .221$ & $.769 \pm .201$ & $.784 \pm .216$ & $.811 \pm .251$ & $.824 \pm .209$ & $.835 \pm .218$ & $.839 \pm .194$ \\
                                            &  \textbf{VStreamDRLS} & $.638 \pm .015$ & $.718 \pm .021$ & $.792 \pm .023$ & $.813 \pm .026$ & $.840 \pm .020$ & $.862 \pm .019$ & $.862 \pm .015$ & $.862 \pm .012$ & $.871 \pm .019$ \\
                                            &  \textbf{MetaHIN}     & $.692 \pm .132$ & $.735 \pm .119$ & $.819 \pm .156$ & $.842 \pm .201$ & $.896 \pm .101$ & $.902 \pm .125$ & $.911 \pm .163$ & $.918 \pm .196$ & $.920 \pm .210$ \\
                                            &  \textbf{PolicyGNN}   & $.649 \pm .025$ & $.682 \pm .017$ & $.701 \pm .019$ & $.724 \pm .020$ & $.731 \pm .015$ & $.739 \pm .018$ & $.752 \pm .012$ & $.781 \pm .017$ & $.812 \pm .029$ \\
                                            &  \textbf{GELS*}       & $.706 \pm .012$ & $.745 \pm .013$ & $.792 \pm .019$ & $.802 \pm .016$ & $.812 \pm .015$ & $.835 \pm .017$ & $.841 \pm .012$ & $.865 \pm .026$ & $.872 \pm .032$ \\
                                            &  \textbf{GELS}        & \textbf{.792} $\pm$ \textbf{.016} & \textbf{.821} $\pm$ \textbf{.018} & \textbf{.845} $\pm$ \textbf{.021} & \textbf{.879} $\pm$ \textbf{.017} & \textbf{.908} $\pm$ \textbf{.012} & \textbf{.918} $\pm$ \textbf{.028} & \textbf{.915} $\pm$ \textbf{.018} & \textbf{.937} $\pm$ \textbf{.019} & \textbf{.945} $\pm$ \textbf{.021} \\
                                             \hline 
        
        \multirow{8}{*}{\textbf{Macro-F1}}  & \textbf{DyGNN}        & $.693 \pm .526$ & $.735 \pm .486$ & $.792 \pm .385$ & $.799 \pm .369$ & $.805 \pm .362$ & $.841 \pm .486$ & $.855 \pm .317$ & $.859 \pm .396$ & $.872 \pm .417$ \\
                                            &  \textbf{EvolveGCN}   & $.752 \pm .173$ & $.796 \pm .178$ & $.821 \pm .196$ & $.842 \pm .192$ & $.861 \pm .203$ & $.877 \pm .184$ & $.892 \pm .192$ & $.913 \pm .126$ & $.925 \pm .172$ \\
                                            &  \textbf{DySAT}       & $.802 \pm .426$ & $.831 \pm .442$ & $.862 \pm .403$ & $.886 \pm .397$ & $.897 \pm .402$ & $.905 \pm .386$ & $.916 \pm .406$ & $.924 \pm .418$ & $.935 \pm .442$ \\
                                            &  \textbf{VStreamDRLS} & $.823 \pm .126$ & $.842 \pm .172$ & $.867 \pm .168$ & $.887 \pm .192$ & $.892 \pm .177$ & $.901 \pm .200$ & $.908 \pm .135$ & $.912 \pm .169$ & $.917 \pm .178$ \\
                                            &  \textbf{MetaHIN}     & $.874 \pm .396$ & $.869 \pm .381$ & $.892 \pm .327$ & $.902 \pm .318$ & $.923 \pm .495$ & $.928 \pm .336$ & $.942 \pm .360$ & $.956 \pm .381$ & $.956 \pm .401$ \\
                                            &  \textbf{PolicyGNN}   & $.852 \pm .161$ & $.861 \pm .172$ & $.877 \pm .129$ & $.889 \pm .166$ & $.902 \pm .169$ & $.915 \pm .195$ & $.921 \pm .114$ & $.931 \pm .186$ & $.935 \pm .103$ \\
                                            &  \textbf{GELS*}       & $.812 \pm .016$ & $.833 \pm .051$ & $.855 \pm .025$ & $.861 \pm .021$ & $.872 \pm .016$ & $.886 \pm .024$ & $.893 \pm .011$ & $.899 \pm .021$ & $.902 \pm .021$ \\
                                            &  \textbf{GELS}        & \textbf{.895} $\pm$ \textbf{.024} & \textbf{.902} $\pm$ \textbf{.038} & \textbf{.916} $\pm$ \textbf{.016} & \textbf{.919} $\pm$ \textbf{.026} & \textbf{.927} $\pm$ \textbf{.025} & \textbf{.942} $\pm$ \textbf{.051} & \textbf{.964} $\pm$ \textbf{.012} & \textbf{.972} $\pm$ \textbf{.014} & \textbf{.982} $\pm$ \textbf{.024} \\
                                             \hline 
        
        \multirow{8}{*}{\textbf{Micro-F1}}  & \textbf{DyGNN}        & $.524 \pm .261$ & $.592 \pm .272$ & $.651 \pm .297$ & $.699 \pm .271$ & $.728 \pm .197$ & $.769 \pm .206$ & $.792 \pm .214$ & $.812 \pm .262$ & $.819 \pm .169$ \\
                                            &  \textbf{EvolveGCN}   & $.689 \pm .191$ & $.693 \pm .216$ & $.722 \pm .204$ & $.776 \pm .196$ & $.792 \pm .173$ & $.809 \pm .188$ & $.815 \pm .196$ & $.843 \pm .159$ & $.870 \pm .198$ \\
                                            &  \textbf{DySAT}       & $.703 \pm .296$ & $.737 \pm .362$ & $.781 \pm .395$ & $.802 \pm .379$ & $.825 \pm .271$ & $.839 \pm .286$ & $.842 \pm .302$ & $.855 \pm .216$ & $.867 \pm .288$ \\
                                            &  \textbf{VStreamDRLS} & $.734 \pm .021$ & $.758 \pm .026$ & $.792 \pm .031$ & $.815 \pm .035$ & $.853 \pm .032$ & $.861 \pm .029$ & $.877 \pm .028$ & $.886 \pm .017$ & $.889 \pm .301$ \\
                                            &  \textbf{MetaHIN}     & $.762 \pm .106$ & $.781 \pm .123$ & $.803 \pm .196$ & $.829 \pm .182$ & $.864 \pm .174$ & $.879 \pm .182$ & $.901 \pm .127$ & $.922 \pm .129$ & $.929 \pm .139$ \\
                                            &  \textbf{PolicyGNN}   & $.741 \pm .021$ & $.775 \pm .026$ & $.800 \pm .031$ & $.816 \pm .018$ & $.859 \pm .027$ & $.871 \pm .031$ & $.889 \pm .102$ & $.901 \pm .025$ & $.901 \pm .038$ \\
                                            &  \textbf{GELS*}       & $.781 \pm .017$ & $.802 \pm .023$ & $.812 \pm .029$ & $.819 \pm .015$ & $.842 \pm .018$ & $.859 \pm .010$ & $.868 \pm .018$ & $.891 \pm .016$ & $.902 \pm .020$ \\
                                            &  \textbf{GELS}        & \textbf{.803} $\pm$ \textbf{.010} & \textbf{.869} $\pm$ \textbf{.019} & \textbf{.882} $\pm$ \textbf{.016} & \textbf{.895} $\pm$ \textbf{.028} & \textbf{.904} $\pm$ \textbf{.016} & \textbf{.921} $\pm$ \textbf{.018} & \textbf{.936} $\pm$ \textbf{.027} & \textbf{.948} $\pm$ \textbf{.031} & \textbf{.951} $\pm$ \textbf{.017} \\
                                             \hline 
        \textbf{LiveStream-2} \\ \hline
                                             
        \multirow{8}{*}{\textbf{AUC}}       & \textbf{DyGNN}        & $.516 \pm .409$ & $.592 \pm .382$ & $.679 \pm .359$ & $.781 \pm .362$ & $.832 \pm .357$ & $.856 \pm .402$ & $.862 \pm .391$ & $.869 \pm .369$ & $.882 \pm .412$ \\
                                            &  \textbf{EvolveGCN}   & $.552 \pm .168$ & $.603 \pm .130$ & $.682 \pm .127$ & $.799 \pm .143$ & $.847 \pm .125$ & $.863 \pm .172$ & $.899 \pm .181$ & $.903 \pm .119$ & $.905 \pm .120$ \\
                                            &  \textbf{DySAT}       & $.567 \pm .201$ & $.618 \pm .205$ & $.691 \pm .206$ & $.803 \pm .212$ & $.852 \pm .210$ & $.877 \pm .201$ & $.916 \pm .211$ & $.919 \pm .215$ & $.920 \pm .218$ \\
                                            &  \textbf{VStreamDRLS} & $.579 \pm .012$ & $.629 \pm .016$ & $.711 \pm .017$ & $.821 \pm .019$ & $.869 \pm .023$ & $.892 \pm .019$ & $.931 \pm .015$ & $.933 \pm .013$ & $.935 \pm .014$ \\
                                            &  \textbf{MetaHIN}     & $.726 \pm .129$ & $.794 \pm .115$ & $.829 \pm .126$ & $.884 \pm .125$ & $.895 \pm .123$ & $.921 \pm .129$ & $.956 \pm .117$ & $.957 \pm .130$ & $.958 \pm .125$ \\
                                            &  \textbf{PolicyGNN}   & $.618 \pm .013$ & $.702 \pm .015$ & $.756 \pm .020$ & $.842 \pm .011$ & $.882 \pm .017$ & $.905 \pm .019$ & $.940 \pm .015$ & $.946 \pm .012$ & $.949 \pm .020$ \\
                                            &  \textbf{GELS*}       & $.602 \pm .018$ & $.682 \pm .021$ & $.712 \pm .019$ & $.765 \pm .021$ & $.792 \pm .020$ & $.812 \pm .018$ & $.844 \pm .017$ & $.879 \pm .018$ & $.899 \pm .016$ \\
                                            &  \textbf{GELS}        & \textbf{.782} $\pm$ \textbf{.025} & \textbf{.836}  $\pm$ \textbf{.023} & \textbf{.869}  $\pm$  \textbf{.022} & \textbf{.902}  $\pm$ \textbf{.026} & \textbf{.923}  $\pm$  \textbf{.023} & \textbf{.947}  $\pm$ \textbf{.028} & \textbf{.968}  $\pm$ \textbf{.024} & \textbf{.969}  $\pm$ \textbf{.021} & \textbf{.971}  $\pm$ \textbf{.019} \\
                                            \hline 
        \multirow{8}{*}{\textbf{Macro-F1}}  & \textbf{DyGNN}        & $.684 \pm .348$ & $.693 \pm .362$ & $.718 \pm .304$ & $.749 \pm .297$ & $.766 \pm .351$ & $.818 \pm .363$ & $.839 \pm .375$ & $.851 \pm .302$ & $.874 \pm .286$ \\
                                            &  \textbf{EvolveGCN}   & $.691 \pm .162$ & $.729 \pm .159$ & $.736 \pm .147$ & $.756 \pm .148$ & $.767 \pm .152$ & $.820 \pm .155$ & $.845 \pm .157$ & $.859 \pm .160$ & $.886 \pm .159$ \\
                                            &  \textbf{DySAT}       & $.718 \pm .392$ & $.733 \pm .389$ & $.745 \pm .386$ & $.768 \pm .397$ & $.773 \pm .401$ & $.794 \pm .388$ & $.837 \pm .402$ & $.855 \pm .399$ & $.874 \pm .386$ \\
                                            &  \textbf{VStreamDRLS} & $.739 \pm .102$ & $.756 \pm .138$ & $.782 \pm .121$ & $.821 \pm .129$ & $.825 \pm .118$ & $.849 \pm .114$ & $.850 \pm .115$ & $.867 \pm .163$ & $.882 \pm .147$ \\
                                            &  \textbf{MetaHIN}     & $.769 \pm .305$ & $.781 \pm .321$ & $.802 \pm .337$ & $.844 \pm .350$ & $.859 \pm .306$ & $.862 \pm .316$ & $.871 \pm .318$ & $.880 \pm .326$ & $.894 \pm .329$ \\
                                            &  \textbf{PolicyGNN}   & $.742 \pm .023$ & $.768 \pm .027$ & $.793 \pm .021$ & $.827 \pm .029$ & $.833 \pm .022$ & $.856 \pm .026$ & $.862 \pm .019$ & $.873 \pm .017$ & $.890 \pm .014$ \\
                                            &  \textbf{GELS*}       & $.729 \pm .017$ & $.774 \pm .014$ & $.803 \pm .018$ & $.831 \pm .020$ & $.857 \pm .017$ & $.879 \pm .016$ & $.884 \pm .014$ & $.892 \pm .018$ & $.903 \pm .019$ \\
                                            &  \textbf{GELS}        & \textbf{.801}  $\pm$ \textbf{.012} & \textbf{.839}  $\pm$ \textbf{.018} & \textbf{.848}  $\pm$ \textbf{.014} & \textbf{.869}  $\pm$ \textbf{.012} & \textbf{.901}  $\pm$  \textbf{.018} & \textbf{.923}  $\pm$ \textbf{.021} & \textbf{.935}  $\pm$  \textbf{.020} & \textbf{.942}  $\pm$ \textbf{.016} & \textbf{.949}  $\pm$ \textbf{.018} \\
                                             \hline 
        \multirow{8}{*}{\textbf{Micro-F1}}  & \textbf{DyGNN}        & $.573 \pm .273$ & $.589 \pm .284$ & $.618 \pm .229$ & $.669 \pm .216$ & $.702 \pm .268$ & $.728 \pm .277$ & $.742 \pm .251$ & $.759 \pm .257$ & $.768 \pm .261$ \\
                                            &  \textbf{EvolveGCN}   & $.586 \pm .183$ & $.592 \pm .126$ & $.621 \pm .179$ & $.674 \pm .127$ & $.706 \pm .196$ & $.743 \pm .152$ & $.758 \pm .163$ & $.771 \pm .159$ & $.782 \pm .127$ \\
                                            &  \textbf{DySAT}       & $.595 \pm .251$ & $.603 \pm .278$ & $.634 \pm .283$ & $.682 \pm .244$ & $.711 \pm .246$ & $.752 \pm .238$ & $.768 \pm .216$ & $.792 \pm .259$ & $.801 \pm .207$ \\
                                            &  \textbf{VStreamDRLS} & $.612 \pm .035$ & $.627 \pm .031$ & $.664 \pm .028$ & $.691 \pm .030$ & $.728 \pm .033$ & $.763 \pm .036$ & $.775 \pm .027$ & $.803 \pm .029$ & $.818 \pm .031$ \\
                                            &  \textbf{MetaHIN}     & $.721 \pm .183$ & $.732 \pm .152$ & $.775 \pm .147$ & $.796 \pm .182$ & $.813 \pm .159$ & $.826 \pm .138$ & $.849 \pm .192$ & $.851 \pm .189$ & $.866 \pm .186$ \\
                                            &  \textbf{PolicyGNN}   & $.698 \pm .032$ & $.712 \pm .036$ & $.743 \pm .038$ & $.758 \pm .030$ & $.782 \pm .035$ & $.791 \pm .032$ & $.816 \pm .036$ & $.825 \pm .036$ & $.843 \pm .035$ \\
                                            &  \textbf{GELS*}       & $.726 \pm .013$ & $.745 \pm .016$ & $.781 \pm .009$ & $.796 \pm .012$ & $.801 \pm .015$ & $.823 \pm .017$ & $.840 \pm .020$ & $.856 \pm .021$ & $.867 \pm .023$ \\
                                            &  \textbf{GELS}        & \textbf{.765} $\pm$ \textbf{.012} & \textbf{.786} $\pm$ \textbf{.015} & \textbf{.803} $\pm$ \textbf{.016} & \textbf{.819} $\pm$ \textbf{.018} & \textbf{.824} $\pm$ \textbf{.013} & \textbf{.843} $\pm$ \textbf{.014} & \textbf{.863} $\pm$ \textbf{.021} & \textbf{.897} $\pm$ \textbf{.025} & \textbf{.904} $\pm$ \textbf{.016} \\
                                            \hline 
        \textbf{LiveStream-3} \\ \hline
        
        \multirow{8}{*}{\textbf{AUC}}       & \textbf{DyGNN}        & $.387 \pm .316$ & $.428 \pm .386$ & $.469 \pm .362$ & $.502 \pm .375$ & $.528 \pm .394$ & $.578 \pm .392$ & $.619 \pm .373$ & $.672 \pm .381$ & $.735 \pm .396$ \\
                                            &  \textbf{EvolveGCN}   & $.412 \pm .111$ & $.436 \pm .116$ & $.472 \pm .117$ & $.513 \pm .128$ & $.536 \pm .123$ & $.582 \pm .119$ & $.632 \pm .112$ & $.689 \pm .116$ & $.748 \pm .117$ \\
                                            &  \textbf{DySAT}       & $.420 \pm .205$ & $.440 \pm .217$ & $.480 \pm .211$ & $.518 \pm .269$ & $.539 \pm .215$ & $.588 \pm .204$ & $.639 \pm .206$ & $.692 \pm .201$ & $.753 \pm .216$ \\
                                            &  \textbf{VStreamDRLS} & $.428 \pm .015$ & $.446 \pm .017$ & $.489 \pm .020$ & $.526 \pm .012$ & $.541 \pm .018$ & $.595 \pm .019$ & $.645 \pm .016$ & $.702 \pm .014$ & $.755 \pm .021$ \\
                                            &  \textbf{MetaHIN}     & $.461 \pm .317$ & $.482 \pm .331$ & $.503 \pm .338$ & $.554 \pm .326$ & $.589 \pm .319$ & $.613 \pm .301$ & $.658 \pm .298$ & $.725 \pm .331$ & $.769 \pm .337$ \\
                                            &  \textbf{PolicyGNN}   & $.401 \pm .021$ & $.436 \pm .022$ & $.492 \pm .023$ & $.512 \pm .024$ & $.534 \pm .015$ & $.602 \pm .018$ & $.642 \pm .016$ & $.711 \pm .020$ & $.761 \pm .021$ \\
                                            &  \textbf{GELS*}       & $.482 \pm .011$ & $.493 \pm .010$ & $.502 \pm .015$ & $.526 \pm .020$ & $.549 \pm .017$ & $.607 \pm .014$ & $.653 \pm .013$ & $.716 \pm .012$ & $.778 \pm .011$ \\
                                            &  \textbf{GELS}        & \textbf{.502} $\pm$ \textbf{.012} & \textbf{.551} $\pm$ \textbf{.016} & \textbf{.568} $\pm$ \textbf{.011} & \textbf{.599} $\pm$ \textbf{.010} & \textbf{.637} $\pm$ \textbf{.015} & \textbf{.648} $\pm$ \textbf{.018} & \textbf{.702} $\pm$ \textbf{.012} & \textbf{.734} $\pm$ \textbf{.017} & \textbf{.782} $\pm$ \textbf{.018} \\
                                             \hline 
        \multirow{8}{*}{\textbf{Macro-F1}}  & \textbf{DyGNN}        & $.459 \pm .297$ & $.467 \pm .265$ & $.512 \pm .278$ & $.558 \pm .241$ & $.603 \pm .286$ & $.649 \pm .257$ & $.683 \pm .231$ & $.704 \pm .259$ & $.738 \pm .244$ \\
                                            &  \textbf{EvolveGCN}   & $.496 \pm .162$ & $.505 \pm .169$ & $.563 \pm .145$ & $.572 \pm .116$ & $.685 \pm .192$ & $.698 \pm .186$ & $.714 \pm .154$ & $.733 \pm .119$ & $.754 \pm .123$ \\
                                            &  \textbf{DySAT}       & $.502 \pm .338$ & $.537 \pm .316$ & $.578 \pm .341$ & $.599 \pm .363$ & $.638 \pm .316$ & $.679 \pm .377$ & $.726 \pm .391$ & $.745 \pm .367$ & $.768 \pm .341$ \\
                                            &  \textbf{VStreamDRLS} & $.528 \pm .017$ & $.544 \pm .012$ & $.591 \pm .014$ & $.614 \pm .016$ & $.642 \pm .018$ & $.686 \pm .014$ & $.730 \pm .011$ & $.756 \pm .020$ & $.771 \pm .013$ \\
                                            &  \textbf{MetaHIN}     & $.582 \pm .316$ & $.602 \pm .382$ & $.624 \pm .301$ & $.639 \pm .313$ & $.668 \pm .297$ & $.692 \pm .278$ & $.742 \pm .316$ & $.768 \pm .381$ & $.792 \pm .316$ \\
                                            &  \textbf{PolicyGNN}   & $.531 \pm .023$ & $.558 \pm .024$ & $.616 \pm .026$ & $.628 \pm .021$ & $.655 \pm .023$ & $.689 \pm .023$ & $.738 \pm .024$ & $.761 \pm .023$ & $.782 \pm .026$ \\
                                            &  \textbf{GELS*}       & $.589 \pm .011$ & $.604 \pm .014$ & $.618 \pm .016$ & $.624 \pm .014$ & $.647 \pm .021$ & $.663 \pm .011$ & $.702 \pm .028$ & $.715 \pm .019$ & $.728 \pm .016$ \\
                                            &  \textbf{GELS}        & \textbf{.616} $\pm$ \textbf{.012} & \textbf{.678} $\pm$ \textbf{.016} & \textbf{.684} $\pm$ \textbf{.011} & \textbf{.703} $\pm$ \textbf{.014} & \textbf{.718} $\pm$ \textbf{.015} & \textbf{.736} $\pm$ \textbf{.012} & \textbf{.769} $\pm$ \textbf{.013} & \textbf{.788} $\pm$ \textbf{.016} & \textbf{.825} $\pm$ \textbf{.014} \\
                                             \hline 
        \multirow{8}{*}{\textbf{Micro-F1}}  & \textbf{DyGNN}        & $.398 \pm .333$ & $.412 \pm .392$ & $.436 \pm .316$ & $.487 \pm .302$ & $.500 \pm .298$ & $.527 \pm .288$ & $.542 \pm .316$ & $.569 \pm .317$ & $.591 \pm .312$ \\
                                            &  \textbf{EvolveGCN}   & $.421 \pm .127$ & $.439 \pm .132$ & $.447 \pm .133$ & $.468 \pm .110$ & $.506 \pm .102$ & $.539 \pm .121$ & $.553 \pm .102$ & $.582 \pm .121$ & $.603 \pm .125$ \\
                                            &  \textbf{DySAT}       & $.479 \pm .216$ & $.492 \pm .212$ & $.538 \pm .283$ & $.559 \pm .231$ & $.612 \pm .244$ & $.628 \pm .247$ & $.637 \pm .219$ & $.641 \pm .211$ & $.655 \pm .268$ \\
                                            &  \textbf{VStreamDRLS} & $.497 \pm .012$ & $.511 \pm .011$ & $.549 \pm .014$ & $.572 \pm .013$ & $.636 \pm .014$ & $.657 \pm .016$ & $.678 \pm .018$ & $.699 \pm .012$ & $.702 \pm .013$ \\
                                            &  \textbf{MetaHIN}     & $.602 \pm .122$ & $.628 \pm .136$ & $.645 \pm .142$ & $.679 \pm .128$ & $.688 \pm .192$ & $.710 \pm .127$ & $.734 \pm .172$ & $.745 \pm .166$ & $.765 \pm .132$ \\
                                            &  \textbf{PolicyGNN}   & $.516 \pm .012$ & $.572 \pm .015$ & $.598 \pm .016$ & $.615 \pm .012$ & $.646 \pm .016$ & $.682 \pm .014$ & $.691 \pm .018$ & $.707 \pm .021$ & $.717 \pm .019$ \\
                                            &  \textbf{GELS*}       & $.623 \pm .014$ & $.689 \pm .011$ & $.702 \pm .012$ & $.738 \pm .013$ & $.749 \pm .014$ & $.761 \pm .013$ & $.774 \pm .010$ & $.764 \pm .012$ & $.792 \pm .013$ \\
                                            &  \textbf{GELS}        & \textbf{.716} $\pm$ \textbf{.013} & \textbf{.737} $\pm$ \textbf{.014} & \textbf{.769} $\pm$ \textbf{.011} & \textbf{.802} $\pm$ \textbf{.015} & \textbf{.813} $\pm$ \textbf{.012} & \textbf{.836} $\pm$ \textbf{.014} & \textbf{.847} $\pm$ \textbf{.013} & \textbf{.862} $\pm$ \textbf{.011} & \textbf{.889} $\pm$ \textbf{.016} \\

    \end{tabular}
    }
    
    \label{tab:comp_over_time}
\end{table*}

In Table \ref{tab:comp_over_time}, we evaluate the performance of the examined models in terms of AUC, Micro-F1, and Macro-F1 over the events' streaming minutes. The proposed GELS model constantly outperforms the baseline approaches in all datasets. This indicates that GELS can efficiently predict the experience of each viewer. Compared with the second best method MetaHIN, GELS achieves relative improvements of $4.67$, $2.16$, and $5.88\%$ in terms of AUC, Macro-F1 and Micro-F1, respectively, in LiveStream-$1$. In the LiveStream-$2$ dataset, the relative improvements are $3.29$, $5.86$, and $3.87\%$, and $7.52$, $6.93$ and $14.43\%$ in LiveStream-$3$. MetaHIN performs better than the other baseline approaches by following a meta-learning strategy on the connections, achieving high prediction accuracy for new viewers that emerge during an event. However, MetaHIN ignores the information provided by historical events and therefore requires a large amount connections to achieve a high classification accuracy. The proposed GELS model overcomes this shortcoming by adopting the boosting strategy on different streaming events, achieving high classification accuracy at the first minutes of the event. We also observe that the GELS* variant has relatively high classification accuracy at the first minutes of the event, when compared with the other baseline strategies. However by not following the gradient boosting strategy, the GELS* variant ignores the extracted information from other events, hence underperforms when compared with the proposed GELS model.

\subsection{Evaluation based on QoE Labels and Connection Quality}
\begin{table}[]
    \centering
    \caption{Models' comparison in terms of average AUC based on the user experience labels.}
    \resizebox{\columnwidth}{!}{
    \begin{tabular}{c|c|c|c|c}
        \textbf{QoE Label} & \textbf{Baselines} & \multicolumn{3}{c}{\textbf{Datasets}}  \\ \hline
         & & \textbf{LiveStream-1} & \textbf{LiveStream-2} & \textbf{LiveStream-3} \\  \hline
         \multirow{8}{*}{\textbf{Bad}}          & \textbf{DyGNN}        & $.417 \pm .327$ & $.369 \pm .294$ & $.441 \pm .286$ \\ 
                                                & \textbf{EvolveGCN}    & $.492 \pm .174$ & $.397 \pm .149$ & $.514 \pm .169$ \\
                                                & \textbf{DySAT}        & $.513 \pm .221$ & $.413 \pm .251$ & $.542 \pm .236$ \\
                                                & \textbf{VStreamDRLS}  & $.557 \pm .017$ & $.458 \pm .014$ & $.601 \pm .011$ \\
                                                & \textbf{MetaHIN}      & $.618 \pm .154$ & $.589 \pm .367$ & $.719 \pm .182$ \\
                                                & \textbf{PolicyGNN}    & $.592 \pm .013$ & $.513 \pm .011$ & $.693 \pm .014$ \\
                                                & \textbf{GELS*}        & $.782 \pm .011$ & $.812 \pm .017$ & $.730 \pm .016$ \\
                                                & \textbf{GELS}         & \textbf{.896} $\pm$ \textbf{.012} & \textbf{.903} $\pm$ \textbf{.013} & \textbf{.782} $\pm$ \textbf{.012} \\
                                                 \hline \hline
                                                
          \multirow{8}{*}{\textbf{Poor}}        & \textbf{DyGNN}        & $.551 \pm .385$ & $.623 \pm .299$ & $.678 \pm .212$ \\ 
                                                & \textbf{EvolveGCN}    & $.562 \pm .168$ & $.654 \pm .172$ & $.689 \pm .129$ \\
                                                & \textbf{DySAT}        & $.589 \pm .228$ & $.682 \pm .216$ & $.692 \pm .241$ \\
                                                & \textbf{VStreamDRLS}  & $.612 \pm .013$ & $.694 \pm .014$ & $.703 \pm .013$ \\
                                                & \textbf{MetaHIN}      & $.745 \pm .192$ & $.731 \pm .351$ & $.712 \pm .342$ \\
                                                & \textbf{PolicyGNN}    & $.702 \pm .016$ & $.708 \pm .018$ & $.706 \pm .015$ \\
                                                & \textbf{GELS*}        & $.815 \pm .013$ & $.801 \pm .013$ & $.772 \pm .014$ \\
                                                & \textbf{GELS}         & \textbf{.901} $\pm$ \textbf{.014} & \textbf{.932} $\pm$ \textbf{.014} & \textbf{.780} $\pm$ \textbf{.014} \\
                                                \hline \hline
                                                
          \multirow{8}{*}{\textbf{Fair}}        & \textbf{DyGNN}        & $.679 \pm .312$ & $.731 \pm .299$ & $.658 \pm .336$ \\ 
                                                & \textbf{EvolveGCN}    & $.656 \pm .167$ & $.742 \pm .169$ & $.664 \pm .166$ \\
                                                & \textbf{DySAT}        & $.692 \pm .251$ & $.761 \pm .247$ & $.679 \pm .267$ \\
                                                & \textbf{VStreamDRLS}  & $.713 \pm .013$ & $.788 \pm .021$ & $.684 \pm .022$ \\
                                                & \textbf{MetaHIN}      & $.814 \pm .162$ & $.843 \pm .377$ & $.738 \pm .183$ \\
                                                & \textbf{PolicyGNN}    & $.732 \pm .011$ & $.795 \pm .012$ & $.692 \pm .011$ \\
                                                & \textbf{GELS*}        & $.819 \pm .012$ & $.800 \pm .013$ & $.714 \pm .014$ \\
                                                & \textbf{GELS}         & \textbf{.944} $\pm$ \textbf{.017} & \textbf{.936} $\pm$ \textbf{.017} & \textbf{.782} $\pm$ \textbf{.018} 
                                                 \\\hline \hline
                                                
          \multirow{8}{*}{\textbf{Good}}        & \textbf{DyGNN}        & $.703 \pm .302$ & $.713 \pm .284$ & $.652 \pm .291$ \\ 
                                                & \textbf{EvolveGCN}    & $.712 \pm .183$ & $.758 \pm .126$ & $.689 \pm .192$ \\
                                                & \textbf{DySAT}        & $.759 \pm .255$ & $.789 \pm .271$ & $.693 \pm .288$ \\
                                                & \textbf{VStreamDRLS}  & $.812 \pm .014$ & $.802 \pm .015$ & $.712 \pm .015$ \\
                                                & \textbf{MetaHIN}      & $.866 \pm .156$ & $.896 \pm .316$ & $.760 \pm .121$ \\
                                                & \textbf{PolicyGNN}    & $.789 \pm .014$ & $.773 \pm .014$ & $.717 \pm .013$ \\
                                                & \textbf{GELS*}        & $.820 \pm .017$ & $.816 \pm .011$ & $.742 \pm .012$ \\
                                                & \textbf{GELS}         & \textbf{.945} $\pm$ \textbf{.015} & \textbf{.941} $\pm$ \textbf{.015} & \textbf{.782} $\pm$ \textbf{.013} \\
                                                 \hline \hline
                                                
          \multirow{8}{*}{\textbf{Excellent}}    & \textbf{DyGNN}       & $.718 \pm .321$ & $.772 \pm .293$ & $.631 \pm .264$ \\ 
                                                & \textbf{EvolveGCN}    & $.802 \pm .145$ & $.812 \pm .162$ & $.655 \pm .177$ \\
                                                & \textbf{DySAT}        & $.823 \pm .211$ & $.834 \pm .276$ & $.682 \pm .288$ \\
                                                & \textbf{VStreamDRLS}  & $.844 \pm .018$ & $.847 \pm .020$ & $.719 \pm .019$ \\
                                                & \textbf{MetaHIN}      & $.902 \pm .154$ & $.895 \pm .361$ & $.768 \pm .172$ \\
                                                & \textbf{PolicyGNN}    & $.817 \pm .014$ & $.782 \pm .013$ & $.760 \pm .013$ \\
                                                & \textbf{GELS*}        & $.823 \pm .017$ & $.894 \pm .022$ & $.770 \pm .020$ \\
                                                & \textbf{GELS}         & \textbf{.946} $\pm$ \textbf{.016} & \textbf{.950} $\pm$ \textbf{.018} & \textbf{.783} $\pm$ \textbf{.016} \\
         
    \end{tabular}
    }
    
    \label{tab:auc_qoe}
\end{table}

In Table \ref{tab:auc_qoe}, we evaluate the classification accuracy of the examined models for each user experience label, by averaging the AUC metric over all the streaming minutes. We observe that the proposed GELS model achieves high classification accuracy in all user experience labels for all datasets. This indicates the ability of GELS to accurately classify each viewer in the correct user experience label. When compared with the second best approach MetaHIN, the proposed GELS model achieves relative improvements of $31.03$, $17.31$, $13.77$, $8.36$, and $4.65\%$ for LiveStream-$1$  in the bad, poor, fair, good and excellent user experience labels, respectively, for LiveStream-$2$  $34.77$, $21.57$, $9.93$, $4.78$, and $5.70\%$, while for LiveStream-$3$ $8.05$, $8.72$, $5.63$, $2.81$, and $1.92\%$. On inspection of Table \ref{tab:auc_qoe}, we observe that MetaHIN has higher classification accuracy on the fair, good and excellent user experiences than in the bad and poor labels. This means that MetaHIN is biased on the training data and cannot generalize well on all the user experience labels. As aforementioned in Section 1, the main challenge in live video streaming events is to predict the bad and poor labels, since the majority of the viewers have poor user experience at the first minutes of the event. The proposed GELS model handles this problem by employing a replay memory buffer to store the most diverse state-action transitions in terms of the viewers' user experience.

In Table \ref{tab:auc_connection}, we study the impact of the connection level on the classification accuracy of the examined models in terms of average AUC. We observe that the proposed GELS model bests the baseline strategies for all connection levels. This occurs because our model exploits information from several past live video streaming events. In doing so, the learned policy accurately predicts the user experience level of each viewer, regardless of the network capacity of their connections. When compared with MetaHIN, the proposed GELS model achieves relative improvements $28.62$, $13.11$, and $4.44\%$ in  LiveStream-$1$  in the low, medium, and high connection levels, respectively. Similarly, the relative improvements in Livestream-$2$ are $44.27$, $13.56$, and $6.22\%$, and $26.04$, $4.62$, $1.41\%$ for the LiveStream-$3$ dataset. As aforementioned in Section \ref{sec:data}, most of the viewers establish low bandwidth connections at the beginning of the event, hence, it is essential to accurately classify the low, and medium connection levels. In doing so, GELS can improve the user experience of the viewers with low-bandwidth connections at the first streaming minutes of the live video events.

\begin{table}[t]
    \centering
    \caption{Impact of the connection level on average AUC.}
    \resizebox{\columnwidth}{!}{
    \begin{tabular}{c|c|c|c|c}
        \textbf{Connection Level} & \textbf{Baselines} & \multicolumn{3}{c}{\textbf{Datasets}}  \\ \hline
         & & \textbf{LiveStream-1} & \textbf{LiveStream-2} & \textbf{LiveStream-3} \\  \hline
         \multirow{8}{*}{\textbf{Low}}          & \textbf{DyGNN}        & $.241 \pm .352$ & $.252 \pm .285$ & $.392 \pm .299$ \\ 
                                                & \textbf{EvolveGCN}    & $.255 \pm .170$ & $.314 \pm .155$ & $.402 \pm .136$ \\
                                                & \textbf{DySAT}        & $.318 \pm .242$ & $.388 \pm .275$ & $.431 \pm .281$ \\
                                                & \textbf{VStreamDRLS}  & $.482 \pm .015$ & $.396 \pm .019$ & $.452 \pm .016$ \\
                                                & \textbf{MetaHIN}      & $.671 \pm .162$ & $.521 \pm .366$ & $.551 \pm .128$ \\
                                                & \textbf{PolicyGNN}    & $.582 \pm .013$ & $.412 \pm .014$ & $.482 \pm .013$ \\
                                                & \textbf{GELS*}        & $.813 \pm .014$ & $.821 \pm .013$ & $.682 \pm .014$ \\
                                                & \textbf{GELS}         & \textbf{.940} $\pm$ \textbf{.012} & \textbf{.935} $\pm$ \textbf{.012} & \textbf{.745} $\pm$ \textbf{.010} \\
                                                 \hline \hline
          
          \multirow{8}{*}{\textbf{Medium}}        & \textbf{DyGNN}        & $.681 \pm .316$ & $.574 \pm .261$ & $.601 \pm .257$ \\ 
                                                & \textbf{EvolveGCN}    & $.702 \pm .155$ & $.612 \pm .182$ & $.623 \pm .193$ \\
                                                & \textbf{DySAT}        & $.733 \pm .261$ & $.633 \pm .298$ & $.656 \pm .291$ \\
                                                & \textbf{VStreamDRLS}  & $.742 \pm .014$ & $.689 \pm .014$ & $.681 \pm .012$ \\
                                                & \textbf{MetaHIN}      & $.815 \pm .193$ & $.803 \pm .361$ & $.743 \pm .162$ \\
                                                & \textbf{PolicyGNN}    & $.792 \pm .024$ & $.725 \pm .013$ & $.692 \pm .014$ \\
                                                & \textbf{GELS*}        & $.822 \pm .013$ & $.852 \pm .014$ & $.702 \pm .012$ \\
                                                & \textbf{GELS}         & \textbf{.938} $\pm$ \textbf{.012} & \textbf{.929} $\pm$ \textbf{.014} & \textbf{.779} $\pm$ \textbf{.013} \\
                                                 \hline \hline
          
          \multirow{8}{*}{\textbf{High}}    & \textbf{DyGNN}       & $.732 \pm .327$ & $.775 \pm .270$ & $.629 \pm .216$ \\ 
                                                & \textbf{EvolveGCN}    & $.805 \pm .136$ & $.816 \pm .138$ & $.678 \pm .192$ \\
                                                & \textbf{DySAT}        & $.825 \pm .277$ & $.831 \pm .215$ & $.689 \pm .271$ \\
                                                & \textbf{VStreamDRLS}  & $.841 \pm .012$ & $.842 \pm .013$ & $.710 \pm .013$ \\
                                                & \textbf{MetaHIN}      & $.905 \pm .186$ & $.890 \pm .322$ & $.770 \pm .195$ \\
                                                & \textbf{PolicyGNN}    & $.813 \pm .012$ & $.780 \pm .021$ & $.754 \pm .020$ \\
                                                & \textbf{GELS*}        & $.883 \pm .016$ & $.892 \pm .017$ & $.773 \pm .015$ \\
                                                & \textbf{GELS}         & \textbf{.947} $\pm$ \textbf{.017} & \textbf{.949} $\pm$ \textbf{.017} & \textbf{.781} $\pm$ \textbf{.018} \\

    \end{tabular}
    }
    
    \label{tab:auc_connection}
\end{table}

\begin{figure*}[t]
    \centering
    \includegraphics[scale=0.24]{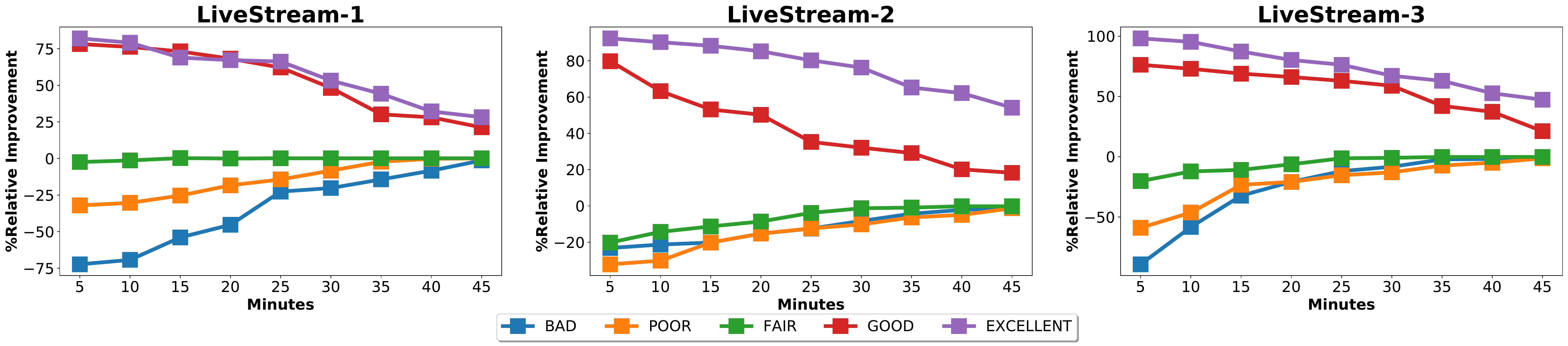}
    \caption{Relative improvement of the GELS model in terms of \#viewers for each user experience level, compared to the initial experience labels of Figure \ref{fig:qoe_viewers}. GELS significantly increases the \#viewers with the good and excellent experience labels at the first streaming minutes.}
    \label{fig:improvement}
\end{figure*}
\subsection{Improvement of User Experience}
So far, in the previous sets of experiments in Sections 4.3 and 4.4 we evaluated the prediction accuracy of the user experience labels. A pressing question resides on how well the proposed GELS model can not only accurately predict the user experience label, but also improve the user experience in the next streaming minutes by establishing the right connections between viewers. In this set of experiments, we measure the impact of the proposed GELS model, comparing the initial user experience labels of Figure \ref{fig:qoe_viewers} with the user experience labels when we apply GELS and take the necessary actions to improve the user experience as presented in Section 2.2. In Figure \ref{fig:improvement}, we present the relative improvement of the proposed GELS model in terms of average number of viewers, categorized by the user experience labels over the streaming minutes. We observe that in all datasets our model achieves more than $75\%$ relative improvement on the viewers that achieve good and excellent user experience at the first $5$ minutes of the events. Note that, as described in Section \ref{sec:data}, in the LiveStream-$1$ and LiveStream-$3$ datasets, initially most of the viewers achieve good or excellent user experience after the first $20$ minutes of the event. By employing our model the tracker can predict the user experience of each viewer, and furthermore incorporate the predicted user experience in the connection selection process of the tracker. Thus, our model can clearly improve the user experience for most of the viewers at the first minutes of the events.

\section{Conclusions} \label{sec:conclusions}

In this paper, we presented a deep graph reinforcement learning model for improving the experience of each viewer when attending enterprise live video streaming events. In the proposed GELS model we formulate each event as a MDP and address the problem of viewers' connection selection based on a graph reinforcement learning strategy. Our model focuses on the viewers that exhibit poor user experience at the first minutes and are connected with low-bandwidth connections. To improve the user experience of the viewers at the first minutes of the event, our model is trained on diverse state-action transitions based on the KL-divergence of each viewer's experience. In addition, we employ a gradient boosting strategy, to generate a global model that learns from different events. In doing so, our model achieves high prediction accuracy on the user experience at the first minutes of an event, and incorporates the predictions in the connections selection process of the tracker. Our experiments on three real-world datasets showed that the proposed GELS model achieves high classification accuracy, with average relative improvements of $5.16$, $4.98$, $8.72\%$ in terms of AUC, Macro-F1 and Micro-F1, respectively, when compared with the second best strategy.

The proposed GELS model can be directly deployed to real-world live video events and improve the user experience over the streaming minutes. Over the last  years, there has been a remarkable demand of  live video streaming events by large enterprises. Since $2019$, Fortune-500 companies host more than $93\%$  events every year for both external and internal communication \cite{conviva}. This growth occurs because viewers retain $95\%$ of a message when they watch it in a video content, while only $10\%$ when the same message is shared through text \cite{Lin2011video,popvideo}. On a daily basis, more than $70\%$ of the employees attend at least one live video streaming \cite{iab}. In doing so, large enterprises are able to improve their employees engagement, and thus their productivity \cite{hiveengagement, microsoft}. However, viewers might lose their interests of the live video streaming event when they perceive poor user experience, and more than $50\%$ of viewers abandon the event before it finishes \cite{akamai}. By deploying our model, large enterprises can attract a large audience to their events and increase their employees' engagement.\cite{dobrian2011understanding}. Large enterprises invest a significant amount of their budget in live video streaming solutions. In $2018$, the enterprise video market was estimated at $\$14$ billion, and is expected to exceed $\$24$ billion by $2026$ \cite{streamingmedia}. This indicates the importance of each live video streaming event to have a high return of investment. This means that the organizations can increase the reach and impact of the video streaming events on most of the employees. Therefore, by employing our model large enterprises can efficiently distribute the video content to all employees at a high-quality, improving significantly the return of investment of each event. As future work we plan to design a multi-task reinforcement learning strategy, to incorporate the video player's bitrate selection process in the live video streaming events \cite{yousef2020adaptive,Wang2020multilive, Meng2019fastconv}.


\balance
\bibliographystyle{ACM-Reference-Format}
\bibliography{sample-base}

\end{document}